%% file: main_arxiv.tex
\documentclass[preprint,12pt]{article}

\usepackage{geometry}

\usepackage{amsmath}
\usepackage{amssymb}
\usepackage{bm}

\usepackage{booktabs}
\usepackage{array}
\usepackage{multirow}
\usepackage{makecell}
\usepackage{threeparttable}
\usepackage{arydshln}
\usepackage{adjustbox}

\usepackage{graphicx}
\usepackage{float}
\usepackage{stfloats}
\usepackage{afterpage}

\usepackage[title]{appendix}

\usepackage[caption=false,font=footnotesize]{subfig}

\usepackage{tikz}
\usetikzlibrary{
    bayesnet,
    positioning,
    arrows.meta,
    fit,
    backgrounds,
    shapes.geometric,
    calc
}

\usepackage[table]{xcolor}
\usepackage[hidelinks]{hyperref}

\definecolor{azulUC3M}{RGB}{0,0,102}

\newcommand{\alex}[1]{{\color{black}{#1}}}
\newcommand{\carlos}[1]{{\color{black}{#1}}}

\usepackage{verbatim}
\usepackage{pdfpages}

\providecommand{\keywords}[1]{\textbf{\textit{Keywords---}} #1}

\date{}
\begin{document}

\title{Biologically Informed Representation Learning for Robust Cross-Center Generalization of MALDI-TOF Mass Spectrometry}

\author{
  Alejandro L. García-Navarro$^{1,2,}$\thanks{Corresponding author (e-mail: agnavarr@pa.uc3m.es)} \and
  Carlos Sevilla-Salcedo$^{1}$ \and
  Belén Rodríguez-Sánchez$^{2,3}$ \and
  Vanessa G\'{o}mez-Verdejo$^{1,2}$
}

\date{
\footnotesize
  $^{1}$ Department of Signal Theory and Communications, Universidad Carlos III de Madrid, Leganés, 28911, Spain \\
  $^{2}$ Instituto de Investigación Sanitaria Gregorio Marañón (IiSGM), Madrid, 28009 Spain \\
  $^{3}$ Clinical Microbiology and Infectious Diseases Department at Hospital General Universitario Gregorio Marañón, Madrid, 28007, Spain
}

\maketitle

\begin{abstract}

Machine learning models for MALDI-TOF mass spectrometry have shown considerable promise for clinical microbiology tasks such as microbial identification and antimicrobial resistance prediction. However, their deployment across institutions remains limited by domain shift, as acquisition-specific variability often leads models to capture technical artifacts rather than transferable biological information. Existing representation learning approaches primarily address this problem through statistical domain alignment while largely overlooking the biological supervision naturally available in microbiology datasets.

We introduce DALMA, a probabilistic representation learning framework that jointly models acquisition-specific variability and biological supervision to learn biologically structured latent representations. By combining domain-specific reconstruction with biologically guided representation learning, DALMA learns transferable representations that generalize across heterogeneous clinical centers without requiring institution-specific components at inference, enabling zero-shot deployment on previously unseen sites.



We evaluate DALMA on a multi-center benchmark comprising seven datasets from three countries. DALMA consistently achieves state-of-the-art zero-shot microbial identification across two held-out clinical centers, while the learned representations also transfer effectively to antimicrobial resistance prediction. Furthermore, latent-space novelty estimation enables reliable selective prediction under previously unseen domain shifts. These results demonstrate that biologically informed representation learning provides an effective strategy for robust and transferable ML in clinical microbiology.
\end{abstract}


\input{Inputs/intro_jbhi}

\section{Materials}
\label{materials}

\input{Inputs/materials_jbhi}

\section{Methodology}
\label{sec:methodology}

\input{Inputs/methods_jbhi}

\section{Experimental Setup}
\label{sec:experimental-setup}
\input{Inputs/experimental_setup_2}

\section{Results}
\label{sec:results}

\input{Inputs/results_jbhi}


\section{Conclusions}
\input{Inputs/conclusions_jbhi}

\section*{Acknowledgements}
CSS and VGV acknowledge financial support from grant TEC-2024/COM-89 funded by the Autonomous Community of Madrid. CSS is also supported by the Comunidad de Madrid through the 2025 César Nombela programme (Grant 2025-T1/COM-36091) and by grant PID2025-174087OA-I00 funded by MCIN/AEI/10.13039/501100011033 and ERDF/EU. VGV is partially supported by grant PID2023-146684NB-I00 funded by MCIN/AEI/10.13039/501100011033 and ERDF/EU.

\bibliographystyle{IEEEtran}
\bibliography{references}

\newpage
\begin{appendices}
\input{appendix}
\end{appendices}

\end{document}

%% file: Inputs/intro_jbhi.tex
\section{Introduction}
\label{sec:introduction}

Precise and timely identification of microorganisms is fundamental to clinical microbiology, as it directly guides antimicrobial treatment decisions and shapes infection control strategies~\cite{zukowska2021advanced}. Delays in species characterisation often force clinicians to rely on broad-spectrum treatments, increasing patient risk and promoting the emergence and spread of antimicrobial resistance (AMR)~\cite{weis2022driams}. Consequently, technologies capable of providing reliable microbial profiling within clinically relevant timeframes are essential to improve patient care.

Matrix-Assisted Laser Desorption/Ionisation Time-of-Flight Mass Spectrometry (MALDI-TOF MS) has transformed this process into fast and low-cost species identification directly from microbial colonies~\cite{weis2020review,mortier2021,croxatto2012}. The technique generates a proteomic fingerprint that reflects the abundance of highly expressed proteins across the mass-to-charge ($m/z$) spectrum and has become the standard identification platform in clinical microbiology laboratories worldwide. Beyond species identification, recent studies have demonstrated that the same spectra can support more advanced predictive tasks, including AMR prediction, fine-grained taxonomic characterisation, and other clinically relevant downstream applications, opening new possibilities for data-driven clinical microbiology~\cite{schmidt2026review,astudillo2024,dewaele2024multimodal}.

The success of Machine Learning (ML) on MALDI-TOF MS data has further expanded these possibilities. Classical approaches such as Support Vector Machines and Random Forests, as well as more recent deep learning models, have achieved competitive performance across a variety of microbiological tasks~\cite{weis2020review,mortier2021,schmidt2026review}. However, despite these promising results, a critical obstacle remains largely unresolved: models trained on data from one institution often experience substantial performance degradation when deployed at another.

This limitation stems from \emph{domain shift}, caused by systematic non-biological variation from differing data acquisition conditions. In MALDI-TOF MS, spectra are influenced by the biological identity of the isolate as well as by technical and epidemiological factors associated with the acquisition site. Differences in instrumentation, calibration procedures, sample preparation protocols, and preprocessing pipelines, together with epidemiological differences across patient populations, alter the observed spectral distribution~\cite{topic2023sample}. As a result, ML models may inadvertently learn acquisition-specific patterns that correlate with microbial species in the training data but fail to generalise beyond the original institution.

This phenomenon has been repeatedly documented in the literature. 
Previous studies have shown substantial performance degradation when transferring models across hospitals~\cite{weis2022driams,msumg_paper,martinezmanjon2026}, even between institutions using identical MALDI-TOF MS instrumentation and acquisition protocols~\cite{martinezmanjon2026}.
More recently, Chen et al.~\cite{chen2026benchmarking} showed that retraining models using target-domain data remains necessary to recover competitive performance. Together, these studies establish domain shift as one of the principal barriers to the large-scale deployment of ML models in clinical microbiology.

A common strategy for addressing domain shift is representation learning, where models seek latent representations that preserve biologically relevant information while suppressing acquisition-specific variability~\cite{bengio2013representation}. Existing approaches pursue this objective primarily through statistical alignment. Variational Autoencoders (VAEs)~\cite{kingma2013vae} learn compact latent representations but provide no explicit mechanism to separate biological and technical sources of variation. Domain adaptation methods such as Domain-Adversarial Neural Networks (DANN)~\cite{ganin2016} and Correlation Alignment (CORAL)~\cite{sun2016deep} explicitly align feature distributions across domains, while more recent large-scale pretraining approaches, including the Maldi Transformer~\cite{dewaele2025pre}, learn reusable spectral representations from large collections of MALDI-TOF spectra. Despite their methodological differences, these approaches largely treat domain shift as a distribution-matching problem.


Unlike many domain adaptation settings, MALDI-TOF MS datasets naturally provide reliable biological annotations that can guide representation learning. 
Microbial species labels are available for every spectrum and constitute an acquisition-invariant source of supervision, while downstream microbiological tasks such as AMR prediction provide complementary biological information.
Despite their availability, these sources of biological supervision remain largely underexploited. This perspective extends the objective of representation learning beyond removing acquisition-specific variability, emphasising the preservation of biologically meaningful structure despite domain heterogeneity.

\begin{table*}[!ht]
\centering
\footnotesize
\setlength{\tabcolsep}{3pt}
\caption{Summary of the MALDI-TOF MS benchmark. For each acquisition domain, we report the country, institution, acquisition period, instrumentation, and the number of spectra available for each target species.}
\label{tab:datasets}
\adjustbox{width=\textwidth}{
\begin{tabular}{lllll|rrrrrr}
\toprule
Dataset & Country & Institution & Years & Hardware &
\textit{ECC} & \textit{Ecoli} & \textit{Efaecium} & \textit{Kpn} & \textit{Paer} & \textit{Saur} \\
\midrule
DRIAMS-A & Switzerland & Univ. Hospital Basel & 2015--2018 &
Microflex LT/SH + Smart LS &
2356 & 7320 & 1751 & 3921 & 4852 & 6994 \\

DRIAMS-B & Switzerland & Canton Hospital Basel-Land & 2018 &
Microflex LT-SH &
139 & 838 & 71 & 189 & 190 & 528 \\

DRIAMS-C & Switzerland & Canton Hospital Aarau & 2018 &
Microflex LT-SH &
196 & 927 & 93 & 366 & 357 & 738 \\

DRIAMS-D & Switzerland & Viollier & 2018 &
Microflex Smart LS &
437 & 1998 & 171 & 2151 & 344 & 2168 \\

MARISMa & Spain & Hospital G. U. Gregorio Marañón & 2018--2024 &
Microflex LT/SH + Smart LS &
1534 & 21211 & 2716 & 18591 & 14061 & 18059 \\

RKI & Germany & Robert Koch Institute & Various$^{a}$ &
Autoflex$^{b}$ &
46 & 62 & 357 & 11 & 60 & 228 \\

MS-UMG & Germany & Univ. Medical Center Göttingen & 2020--2021 &
Microflex LT-SH + Smart &
1610 & 11041 & 2115 & 3169 & 3647 & 6708 \\
\bottomrule
\end{tabular}
}
\vspace{0.3em}
\footnotesize
$^{a}$The original publication reports data collected over nearly 20 years, although the exact acquisition dates are not specified. 
$^{b}$Instrument family inferred from acquisition metadata; the exact model is not explicitly reported.
\end{table*}

To investigate this hypothesis, we introduce \textit{DALMA} (\textbf{D}omain \textbf{AL}ignment for \textbf{MA}LDI-TOF MS), a probabilistic representation learning framework that combines a shared encoder, domain-specific decoders, a species-conditioned latent prior, and auxiliary biological supervision to integrate complementary sources of biological information into a unified latent representation.
By explicitly disentangling biological information from acquisition-specific variability during representation learning, DALMA learns transferable latent representations that generalize to unseen clinical centers. Importantly, domain information is only required during training. At inference, DALMA reduces to a single shared encoder, enabling straightforward integration into existing MALDI-TOF MS workflows.

The main contributions of this work are as follows:
\begin{itemize}
\item We introduce DALMA, a probabilistic representation learning framework that combines domain-specific reconstruction with biologically structured latent representations to disentangle biological and acquisition-specific sources of variability in MALDI-TOF spectra.
\item We demonstrate, through a large multi-center benchmark, that biologically informed representations substantially improve zero-shot (ZS) transfer across heterogeneous clinical centers.
\item We show that the learned representations support multiple downstream tasks, including microbial identification and transferable AMR prediction, while enabling selective prediction through latent-space novelty estimation.
\end{itemize}

%% file: Inputs/materials_jbhi.tex
To evaluate DALMA under realistic cross-institutional conditions, we assembled a benchmark of seven independent MALDI-TOF MS datasets from three countries, multiple acquisition sites, and heterogeneous preprocessing pipelines.

\subsection{Datasets}
\label{subsec:datasets}

The benchmark comprises seven acquisition domains extracted from four publicly available repositories: DRIAMS \cite{weis2022driams}, MARISMa \cite{marisma}, RKI \cite{rki}, and MS-UMG \cite{msumg_paper} (Table~\ref{tab:datasets}). Together, these datasets capture variability arising from different institutions, acquisition periods, instrumentation, and preprocessing strategies. In particular, DRIAMS contributes four independent hospital cohorts acquired under a common protocol, MARISMa extends the benchmark with data from a Spanish tertiary-care hospital, RKI provides a curated reference collection, and MS-UMG introduces an additional source of variability through an independent preprocessing pipeline.

Experiments focus on six clinically relevant bacterial groups represented across all datasets: \textit{Enterobacter cloacae} complex (ECC), \textit{Escherichia coli} (Ecoli), \textit{Enterococcus faecium} (Efaecium), \textit{Klebsiella pneumoniae} (Kpn), \textit{Pseudomonas aeruginosa} (Paer), and \textit{Staphylococcus aureus} (Saur). Table~\ref{tab:datasets} summarizes the number of spectra available for each species and acquisition domain.



For AMR, resistance annotations were available for the DRIAMS cohorts, MARISMa, and MS-UMG. 
\alex{The evaluation is restricted to \textit{Kpn}, the species with the most complete resistance annotations across all domains and confirmed proteomic AMR biomarkers within the MALDI-TOF MS detection range~\cite{rodrigueztemporal2026kpc}. Species--antibiotic pairs were retained if they contained at least 10 resistant and 10 susceptible isolates in the evaluation cohort and at least 50 isolates per class across a minimum of two acquisition domains, yielding five prediction tasks for the following antibiotics: Imipenem, Meropenem, Ceftazidime, Ciprofloxacin, and Piperacillin-Tazobactam.}

\subsection{Preprocessing}

All spectra were preprocessed using the standardized pipeline in~\cite{schmidt2026review}, including variance stabilization, smoothing, baseline correction, intensity thresholding, trimming to 2,000--20,000~$m/z$, 3-Da binning, and logarithmic scaling, yielding 6,000-dimensional feature vectors.

MS-UMG is publicly available only in preprocessed form and therefore could not be subjected to the same pipeline, making it an additional source of variability. Nevertheless, both preprocessing procedures produce spectra in the same 6,000-dimensional feature space.

%% file: Inputs/methods_jbhi.tex
\subsection{Problem Formulation}
\label{subsec:problem-formulation}

We consider a multi-center MALDI-TOF MS dataset $\{(\mathbf{x}_i,d_i,\alex{\mathbf{y}_i})\}_{i=1}^{N}$, where $N$ denotes the number of spectra. The $i$-th spectrum $\mathbf{x}_i\in [0,1]^{M}$ is represented by $M$ spectral bins after preprocessing, and is associated with an acquisition domain $d_i\in\{1,\ldots,D\}$. 
\carlos{Additionally, each spectrum is accompanied by a task-dependent annotation $\mathbf{y}_i$, corresponding either to a categorical label (e.g., microbial species identity) or a multi-label vector (e.g., a panel of AMR phenotypes), depending on the prediction task under consideration.} Our objective is to learn a latent representation $\mathbf{z}_i\in\mathbb{R}^{L}$, with $L \ll M$, that captures the biologically relevant information from the spectrum while remaining robust to acquisition-specific variability.  

To achieve robust transfer across institutions, the latent space should satisfy two complementary properties:
\begin{itemize}
\item \textbf{Domain Invariance}: Conditioned on the biological information, latent representations should be consistent across acquisition sites. Formally, the conditional latent representation distribution should exhibit minimal dependence on the acquisition domain:
\begin{equation}
p(\mathbf{z}_i \mid y_i, d{=}a) \approx p(\mathbf{z}_i \mid y_i, d{=}b),\, \forall\, a, b \in \{1, \ldots, D\}.
\end{equation}

\item \textbf{Biological Manifold Structure}: 
Latent representations should be organized according to biologically information rather than acquisition variability. Depending on the available supervision, this biological structure can be imposed either through conditioned latent priors for unique labels, such as microbial species, or through auxiliary supervision for multi-label settings, such as AMR.

\end{itemize}

\alex{As described in next subsections,} DALMA addresses these objectives by learning biologically structured latent representations while  modeling acquisition-specific variability during reconstruction. 

\subsection{Domain Invariance}  
\label{subsec:model-architecture}

\begin{figure*}[!t]
\centering
\includegraphics[width=\textwidth]{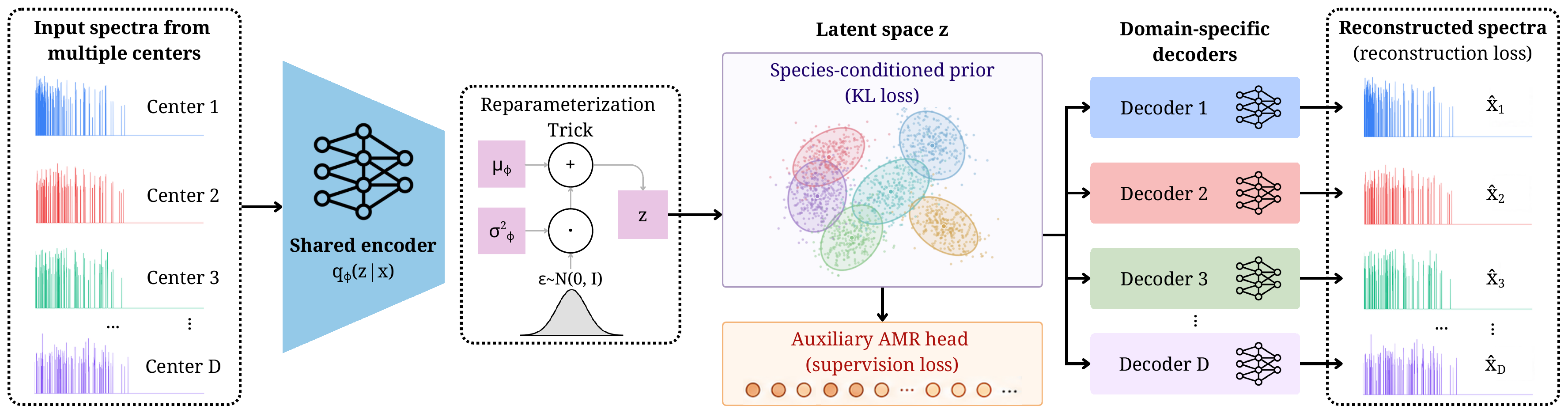}
\caption{Overview of DALMA during training. A shared probabilistic encoder projects spectra from multiple acquisition domains into a common latent representation, while domain-specific decoders reconstruct spectra using acquisition-specific characteristics. Biological supervision is incorporated through a species-conditioned latent prior for microbial identification and, when a unique prior cannot be naturally defined (e.g., AMR), through an auxiliary prediction head jointly optimized during training.}
\label{fig:dalma_arch_1}
\end{figure*}

DALMA models acquisition-specific variability through a shared probabilistic encoder and domain-specific decoders (Fig.~\ref{fig:dalma_arch_1}).
The shared encoder projects spectra acquired at different institutions into a common latent space, while domain-specific decoders model acquisition-dependent variability during reconstruction. This design encourages acquisition-invariant latent representations without requiring explicit domain alignment losses.

To map spectra into a shared latent space, DALMA employs a shared encoder
\begin{equation}
q_{\phi}(\mathbf z \mid \mathbf x)=
\mathcal N
\left(
\mathbf z \mid 
\boldsymbol{\mu}_{\phi}(\mathbf x),
\mathrm{diag}
\left(
\boldsymbol{\sigma}_{\phi}^{2}(\mathbf x)
\right)
\right),
\end{equation}
parameterized by a neural network that predicts the mean and variance of the approximate posterior for each spectrum. Latent samples are obtained using the standard reparameterization trick:
\begin{equation}
\mathbf z = \boldsymbol{\mu}_{\phi}(\mathbf x) +
\boldsymbol{\sigma}_{\phi}(\mathbf x) \odot \boldsymbol{\varepsilon},
\qquad
\boldsymbol{\varepsilon} \sim \mathcal N(\mathbf 0,\mathbf I).
\end{equation}


To reconstruct spectra, DALMA employs a domain-specific decoder for each acquisition domain, $ \left\{
p_{\theta_d}(\mathbf x \mid \mathbf z)
\right\}_{d=1}^{D}$,
such that each decoder models the acquisition-dependent characteristics of its corresponding domain. Let $\textbf{f}_{\theta_d}(\mathbf z)\in [0,1]^M$
denote the output of decoder $d$, and let $f_{\theta_d}^{(m)}(\mathbf z)$
be its $m$-th component. The conditional likelihood is then modeled as
\begin{equation}
p_{\theta_d}(\mathbf{x}\mid\mathbf{z})
=
\prod_{m=1}^{M}
\mathrm{Bernoulli}
\left( x_m \mid f_{\theta_d}^{(m)}(\mathbf z)\right),
\end{equation}
where $x_m$ denotes the intensity of the $m$-th spectral bin.  
By delegating acquisition-specific reconstruction to independent decoders, the shared latent representation is encouraged to capture information that is transferable across acquisition domains while leaving center-dependent effects to the decoder associated with each domain.

\subsection{Biological Supervision}
\label{subsec:biological_supervision}


While the architecture described above models acquisition-specific variability, DALMA explicitly organizes the latent space according to the available biological supervision, whose implementation depends on the downstream task.

For microbial identification, each spectrum is associated with a single species label $s\in\{1,\ldots,S\}$.
This naturally enables defining a species-conditioned latent prior, where each species is represented by its own learnable Gaussian distribution,
\begin{equation}
p(\mathbf z \mid s)=
\mathcal N
\left(
\mathbf z \mid
\boldsymbol{\mu}_s,
\mathrm{diag}
(\boldsymbol{\sigma}_s^2)
\right),
\end{equation}
where $\boldsymbol{\mu}_s,\boldsymbol{\sigma}_s^2\in\mathbb{R}^{L}$ are learnable species-specific parameters. Instead of using a single isotropic Gaussian prior, DALMA computes the KL divergence with respect to the Gaussian associated with the ground-truth species, encouraging spectra from the same microorganism to cluster around a shared latent distribution across acquisition domains.

The resulting objective for microbial identification is therefore given by the modified Evidence Lower Bound (ELBO),
\begin{equation}
\mathcal{L}_{\mathrm{DALMA}} = \mathbb{E}_{q_{\phi}(\mathbf{z}|\mathbf{x})} \left[\log p_{\theta_d}(\mathbf{x}|\mathbf{z})\right] - \mathrm{KL}\left(q_{\phi}(\mathbf{z}|\mathbf{x}) \,\|\, p(\mathbf{z}|s)\right),
\label{eq:dalma_loss}
\end{equation}
where the KL divergence is computed with respect to the species-conditioned prior.


AMR presents a different supervision setting, where each isolate is associated with multiple resistance labels rather than a single biological label. Consequently, a unique latent prior cannot be naturally defined. Instead, DALMA incorporates this biological information through an auxiliary prediction head that guides the latent representation during training. 
This auxiliary head is jointly optimized using a weighted binary cross-entropy loss computed only over samples with available AMR annotations. The overall training objective becomes
\begin{equation}
\mathcal L_{\mathrm{total}}
=
\mathcal L_{\mathrm{DALMA}}
-
\lambda_{\mathrm{AMR}}
\mathcal L_{\mathrm{AMR}},
\label{eq:total_loss}
\end{equation}

where $\mathcal L_{\mathrm{AMR}}$ denotes the weighted binary cross-entropy loss and $\lambda_{\mathrm{AMR}}$ controls the contribution of the auxiliary supervision. For microbial identification experiments, $\lambda_{\mathrm{AMR}}=0$.

\subsection{Downstream Representation}
\label{sec:methods_downstream}

\begin{figure*}[!t]
\centering
\includegraphics[width=\textwidth]{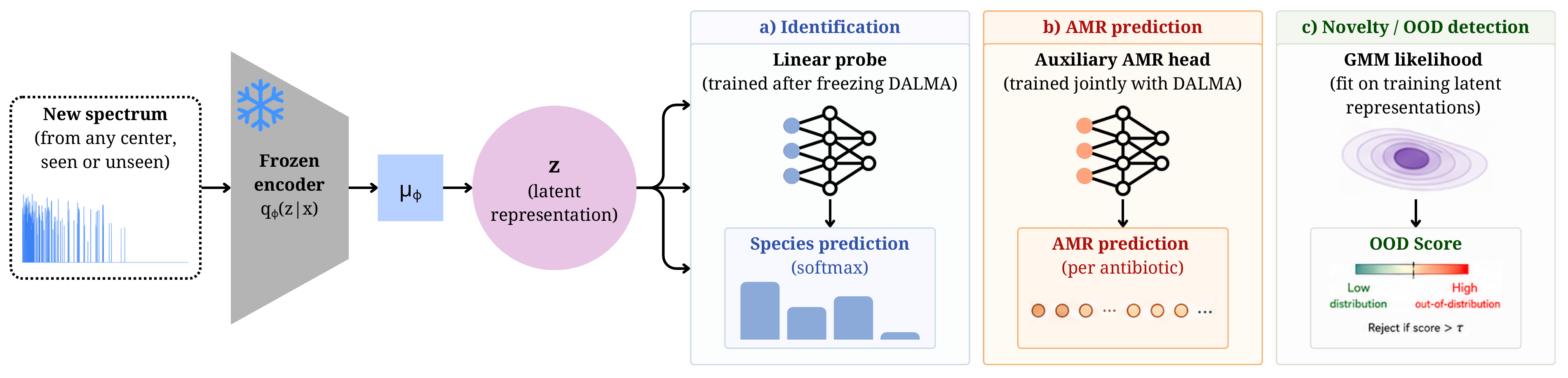}
\caption{Overview of DALMA during inference. After training, only the shared encoder is retained to generate transferable latent representations from unseen spectra. These representations are subsequently used for three downstream tasks: microbial identification using a linear classifier, AMR prediction through the trained auxiliary prediction head, and OOD detection using a GMM fitted to the latent representations of the source domains.}
\label{fig:dalma_arch_2}
\end{figure*}

After training, the goal of DALMA shifts from spectrum reconstruction to representation extraction. Accordingly, only the shared encoder is retained, while the domain-specific decoders are discarded (Fig.~\ref{fig:dalma_arch_2}). Consequently, every spectrum, independently of its acquisition center, is mapped into the common biologically structured latent space through the same encoder.

Rather than sampling from the approximate posterior, downstream tasks use the posterior mean,
\begin{equation}
\mathbf{z}^*=\boldsymbol{\mu}_{\phi}(\mathbf{x}),
\end{equation}
which yields a deterministic spectrum representation   \alex{that preserves} the biological organization learned during training.

These latent representations constitute the input to the downstream analyses considered in this work. Depending on the target application, different lightweight models are built on top of the frozen representation. For microbial identification, a linear classifier \alex{(linear probe)} is trained using the latent embeddings obtained from the source domains. For AMR prediction, the auxiliary prediction head learned during training is directly used for inference. Finally, for out-of-distribution (OOD) detection, a Gaussian Mixture Model (GMM) is fitted to the latent representations of the source training data, allowing the likelihood of unseen samples to be used as a confidence measure.

Therefore, DALMA can be viewed as a transferable representation learning framework in which spectrum reconstruction is only used during training to learn biologically structured latent embeddings, while downstream analyses rely exclusively on the frozen encoder.

%% file: Inputs/experimental_setup_2.tex
\subsection{Source and Target Domains}

All experiments are designed to evaluate the ability of learned representations to generalize across acquisition domains. Hence, we use a multi-source OOD protocol in which the downstream classifier described in Section~\ref{sec:methods_downstream} is trained and validated exclusively on spectra from the source domains (DRIAMS-A, DRIAMS-B, DRIAMS-C, MARISMa, and RKI) and evaluated directly on the held-out target domains (DRIAMS-D and MS-UMG) without any adaptation. 

The two target domains were selected to represent different degrees of distribution shift. DRIAMS-D (Viollier AG, Switzerland) is moderately shifted: although it belongs to the same collection as several source domains, Viollier operates as a diagnostic service provider receiving samples from private practices and hospitals across Switzerland, and spectra are acquired using the Bruker Smart LS platform rather than the Microflex LT-SH systems used at the other DRIAMS sites. These differences were already observed in the original study~\cite{weis2022driams}, 
where DRIAMS-D yielded smaller cross-site performance gains than the remaining domains.



MS-UMG is more challenging, differing simultaneously in instrumentation, geography, acquisition period, and preprocessing pipeline~\cite{msumg_paper}, making it a particularly demanding benchmark for evaluating the robustness of learned representations under severe domain shift.

For each source domain, spectra are partitioned into training, validation, and test subsets using stratified sampling to preserve the species distribution. Unless otherwise specified, 80\% of the spectra are used for training, 10\% for validation, and 10\% as source-domain test data. The validation split is used exclusively for model selection and early stopping. The held-out target domains, DRIAMS-D and MS-UMG, are never used during training or validation.

\subsection{Baseline Methods}
\label{subsec:baselines}

To assess DALMA’s effectiveness, we first include a fully supervised MLP trained directly on the spectra, without an intermediate latent representation. This baseline tests whether domain shift can be mitigated by a sufficiently large classifier on the original feature space, or if explicit representation learning is required.

We then compare DALMA against representative approaches spanning four different strategies for handling domain shift in MALDI-TOF  MS: unsupervised representation learning, adversarial domain adaptation, distribution alignment, and large-scale pretraining.

\begin{itemize}

\item \textbf{Standard VAE}~\cite{kingma2013vae}: A variational autoencoder trained with a standard isotropic Gaussian prior, used as a baseline to test whether latent representation learning alone can mitigate domain shift, without explicit domain alignment or biological structuring.

\item \textbf{DANN} (Domain-Adversarial Neural Network)~\cite{ganin2016}: A domain adaptation approach that encourages domain-invariant representations through adversarial training. A domain classifier is trained jointly with the encoder using a gradient reversal layer, forcing the learned representation to remain predictive of the task while minimizing domain-specific information.

\item \textbf{MultiVAE-CORAL}: A domain-alignment baseline that extends CORAL~\cite{sun2016deep} to the multi-domain setting by minimizing the Frobenius distance between covariance matrices of all source-domain pairs in the latent space, using the same multi-decoder VAE backbone as DALMA.

\item \textbf{Maldi Transformer}~\cite{dewaele2025pre}: A Transformer-based representation learning model for MALDI-TOF spectra that operates on peak-selected spectra and models them as a sequence of informative peaks. 
We evaluate the pretrained XL variant to assess whether large-scale pretraining improves cross-domain generalization.

\end{itemize}

Whenever applicable, baseline models were configured to use latent spaces of the same dimensionality as DALMA to ensure a fair comparison. Detailed architectural specifications and hyperparameter settings for all baselines are provided in the Appendix \ref{app:baselines}.

\subsection{Experimental Protocols}
\label{subsec:experimental_protocols}

\subsubsection{Domain Similarity Analysis}
\label{subsec:domain_similarity}

To quantify the magnitude of the domain shift between acquisition centers, we compare the similarity between domains both in the original spectral space and in the latent representation learned by DALMA, computed on held-out test spectra from the source domains and all available spectra from the target domains. This analysis is intended solely to characterize how the learned representation modifies the geometric relationships between acquisition domains, independently of the downstream classification task.

For two sets of spectra, $X$ and $Y$, we compute their average cosine similarity as
\begin{equation}
\mathrm{sim}(X,Y)=
\frac{1}{|X||Y|}
\sum_{\mathbf{x}\in X}
\sum_{\mathbf{y}\in Y}
\frac{\mathbf{x}^{\top}\mathbf{y}}
{\|\mathbf{x}\|\|\mathbf{y}\|}.
\label{eq:cos_domains}
\end{equation}
To compare acquisition centers, similarities are computed independently for each microbial species and then averaged across species, ensuring that the analysis is not biased by differences in class prevalence between datasets. Let $C_{ij}$ denote the resulting average similarity between domains $i$ and $j$. Since the original spectral space and the DALMA latent space have different dimensionalities and similarity scales, and to ensure a fair comparison, these values are normalized with respect to the within-domain similarities of the corresponding centers,
\begin{equation}
R_{ij}=
\frac{C_{ij}}
{\sqrt{C_{ii}C_{jj}}},  
\label{eq:cos_domain_norm}
\end{equation}
where $R_{ii}=1$ by construction and off-diagonal values closer to one indicate greater similarity between acquisition domains. 

\subsubsection{Species-Conditioned Biological Supervision}
\label{subsec:Exp_protocol_OOD}

The primary objective of this work is to evaluate the ability of the learned representations to transfer across acquisition domains. Following the downstream classification procedure described in Section~\ref{sec:methods_downstream}, the linear probe is trained exclusively on spectra from the source domains and evaluated directly on the held-out target domain without any adaptation. This protocol measures the ability of each representation learning method to support ZS deployment across previously unseen clinical sites.

Additionally, to assess each component of DALMA, we evaluate several ablated variants by independently replacing the species-conditioned prior with a standard isotropic Gaussian prior, and the domain-specific decoders with a single shared decoder, yielding four combinations. All variants are trained under identical conditions and evaluated using the same OOD protocol.

\subsubsection{Auxiliary Biological Supervision}
\label{subsec:amr_protocol}

Unlike microbial identification, AMR prediction is incorporated through an auxiliary prediction head rather than a species-conditioned prior, since each isolate is associated with multiple resistance labels.


Considering \textit{Kpn} and the antibiotics described in Section~\ref{subsec:datasets}, AMR prediction is first evaluated under a ZS cross-domain setting, where the downstream classifier is trained exclusively on source-domain isolates and directly applied to the held-out target domain\footnote{DRIAMS-D is excluded from AMR evaluation since, unlike DRIAMS-A--C, which report R/S/I labels under the EUCAST version current at collection (v6--v8), DRIAMS-D provides raw MIC values converted using EUCAST v9 (2019), released after data collection (2018).}. We then evaluate a few-shot (FS) adaptation scenario by progressively incorporating small numbers of labeled target-domain isolates into the downstream training set while keeping the learned representation frozen. This protocol assesses both the intrinsic transferability of the learned representation and the amount of target-domain supervision required to achieve optimal performance. Each FS setting is repeated over 10 independent random partitions of the target-domain labeled samples.

For this FS evaluation, two baselines are included: an MLP trained on source-domain \textit{Kpn} spectra and fine-tuned on target-domain isolates following the approach of~\cite{chen2026benchmarking}, and a target-only variant trained exclusively on MS-UMG \textit{Kpn} isolates.

\subsubsection{Robustness to Unseen Domain Shifts}
\label{subsubsec:novelty_description}

Although DALMA is designed to reduce acquisition-specific variability, complete robustness to every possible domain shift cannot be guaranteed in real clinical deployment. We therefore evaluate the robustness of the learned representations to unseen domain shifts through an OOD selective prediction protocol based on latent-space novelty detection.

To this end, a GMM is fitted to the latent representations of the training spectra produced by the DALMA encoder. The GMM likelihood is used as a novelty score, where lower likelihood values indicate that a sample lies farther from the high-density regions occupied by the training data.

Novelty detection is evaluated using a selective prediction protocol on the held-out target domains. Spectra with novelty scores below a threshold are rejected, and classification performance is computed on the remaining samples together with the corresponding \emph{coverage}, defined as the proportion of accepted spectra. This evaluates whether the learned latent representation can reliably identify OOD samples.

\subsection{Implementation Details}
\label{subsec:implementation-details}

The encoder and decoder networks were implemented using fully connected neural networks operating on the preprocessed binned spectra. For DALMA, the encoder consists of three hidden layers with $2048$, $1024$, and $512$ units, respectively, followed by two linear projections that output the mean and log-variance of the latent posterior distribution. The latent dimensionality was fixed to $L=64$. 

Each domain-specific decoder mirrors the encoder architecture, using hidden layers of $512$, $1024$, and $2048$ units, followed by a sigmoid output layer of dimension $M$. One decoder was instantiated for each training domain. ReLU activations were used in all hidden layers, while a sigmoid activation was applied at the output to constrain the reconstructed spectrum to the interval $[0,1]$. 
Since input spectra were normalized to the same range, decoder outputs were interpreted as Bernoulli parameters and the reconstruction term was implemented using a Bernoulli likelihood (equivalently, binary cross-entropy reconstruction loss). 

The species-conditional prior was implemented as two learnable parameter matrices of size $S \times L$, storing the mean $\boldsymbol{\mu}_s \in \mathbb{R}^L$ and log-variance $\log\boldsymbol{\sigma}_s^2 \in \mathbb{R}^L$ associated with each species $s$. These parameters are randomly initialized at the start of training and jointly optimized with the encoder and decoders through the ELBO objective in Eq.~\ref{eq:dalma_loss}.  

DALMA was trained using mini-batches of size $128$ and the Adam optimizer \cite{kingma2014} with learning rate $10^{-4}$ and weight decay $10^{-5}$. Early stopping based on validation ELBO was employed with a patience of $20$ epochs. To prevent numerical instabilities from extremely small or large variance estimates, all log-variance parameters
(encoder outputs and species embeddings)
were clamped to the interval $[-6,6]$.


The source code required to reproduce the experiments presented in this work is publicly available on GitHub at \url{https://github.com/alexgaarciia/MALDIAlign}.

\subsection{Evaluation Metrics}

Microbial identification experiments are primarily evaluated using \emph{Balanced Accuracy} (BA), defined as the average recall across microbial species:
\begin{equation}
\text{BA} = \frac{1}{S} \sum_{s=1}^{S} \frac{TP_s}{TP_s + FN_s},
\end{equation}
where $TP_s$ and $FN_s$ denote the number of true positives and false negatives for species $s$, and $S$ is the total number of species. BA is preferred over standard accuracy because species are unevenly represented across datasets, ensuring that minority species contribute equally to the final score.

AMR prediction experiments are evaluated using the area under the receiver operating characteristic curve (AUROC), computed independently for each species--antibiotic prediction task. This metric is appropriate for the class imbalance typically observed in AMR prediction and provides a threshold-independent assessment of discrimination performance.

For the robustness to unseen domain shifts experiments, performance is evaluated using a selective prediction framework. Specifically, we report the BA obtained after rejecting spectra identified as potential outliers together with the corresponding coverage, defined as the proportion of samples for which the model issues a prediction. This allows us to characterize the trade-off between predictive performance and abstention under increasing levels of distribution shift.

Additional metrics, including macro-averaged F1-score, recall, specificity, and AUROC for microbial identification, are provided in the Appendix \ref{app:ood_extended}.

%% file: Inputs/results_jbhi.tex
The experiments are designed to evaluate the proposed latent representations from four complementary perspectives. First, we assess their discriminative quality for microbial identification. We then analyze the contribution of each architectural component through ablation studies, evaluate their ability to generalize across unseen acquisition domains, investigate whether they preserve clinically relevant phenotypic information through AMR prediction, and finally assess the reliability of the learned representations for OOD detection.


\subsection{Learning Domain-Invariant Representations}
\label{subsec:representation_alignment}

\begin{figure}[t!]
\centering
\includegraphics[width=0.9\linewidth]{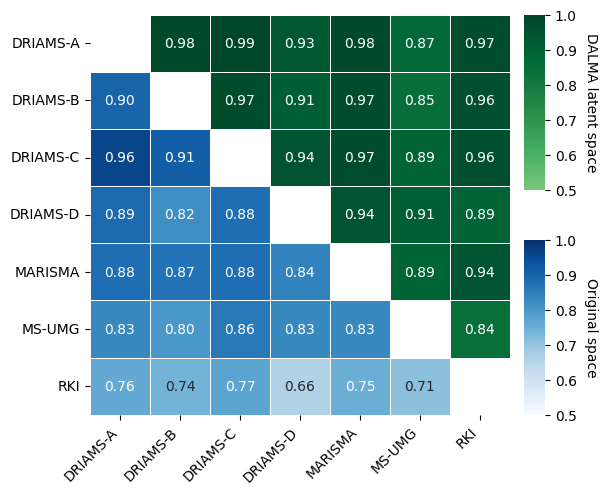}
\caption{Pairwise normalized cosine similarity between acquisition centers, averaged across microbial species. The lower triangle (blue) shows values in the original spectral space; the upper triangle (green) shows values in the DALMA latent space.}
\label{fig:similarity-matrices}
\end{figure}

Since the first objective of DALMA is to learn domain-invariant latent representations, we first evaluate whether the learned embeddings effectively reduce acquisition-specific variability before considering downstream prediction tasks.

Figure~\ref{fig:similarity-matrices} compares the normalized pairwise cosine similarity matrices computed in the original spectral space and in the latent space learned by DALMA, following the protocol described in Section~\ref{subsec:domain_similarity}. Higher values indicate greater similarity between acquisition centers and therefore reduced domain-specific variability.

The original spectral space (blue) shows heterogeneous similarity, with several acquisition centers---particularly DRIAMS-D and RKI—being much less similar to the remaining institutions. In contrast, the latent representations learned by DALMA (green) produce a far more homogeneous similarity pattern across centers. The average normalized cross-domain similarity rises from $0.826$ to $0.931$, while its standard deviation falls from $0.072$ to $0.043$,
indicating that spectra from different acquisition protocols become substantially more consistent in the learned representation.

Importantly, this increased cross-domain similarity does not arise from collapsing spectra into an undifferentiated latent space. Instead, acquisition-specific variability is absorbed by the domain-specific decoders while the shared encoder learns a common latent representation. As shown in the next sections, this representation remains biologically discriminative despite the increased alignment across acquisition domains.


\subsection{Species-Conditioned Biological Supervision}
\label{subsec:species_supervision}
We next evaluate the first biological supervision mechanism proposed in DALMA, namely the species-conditioned latent prior. The objective of these experiments is to assess whether organizing the latent representation according to microbial identity improves the quality and transferability of the learned embeddings. To this end, we evaluate microbial identification under a challenging ZS cross-domain setting and subsequently analyze the contribution of each architectural component through an ablation study.

\subsubsection{Microbial Identification}

To investigate whether the learned latent representations preserve microbial identity across clinical centers, following the protocol described in Section~\ref{subsec:Exp_protocol_OOD}, we froze the shared encoder 
after training and fitted a linear classifier 
on the latent representations extracted from the source domains. Generalization is evaluated under a ZS protocol, where the classifier is tested on acquisition centers that were not observed during representation learning.

\begin{figure}[!t]
\centering
\subfloat[Raw spectra (DRIAMS-D)\label{fig:cm-raw-driamsd}]{
\includegraphics[width=0.5\linewidth]{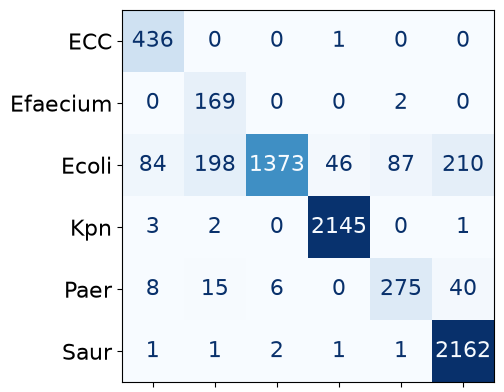}}
\hspace{-0.20cm}
\subfloat[DALMA (DRIAMS-D)\label{fig:cm-dalma-driamsd}]{
\includegraphics[width=0.39\linewidth]{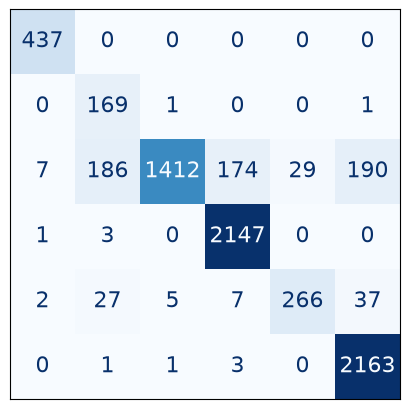}}
\vspace{-1mm}
\subfloat[Raw spectra (MS-UMG)\label{fig:cm-raw-msumg}]{
\includegraphics[width=0.5\linewidth]{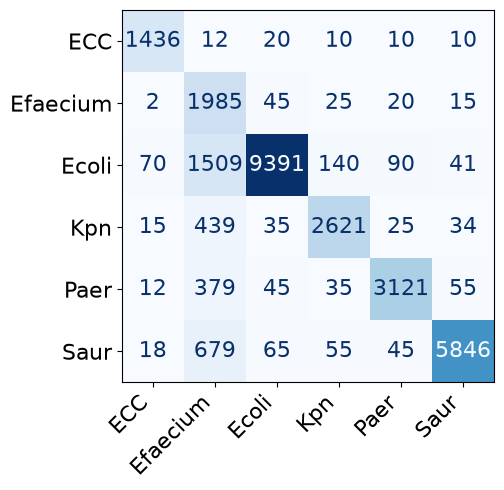}}
\hspace{-0.20cm}
\subfloat[DALMA (MS-UMG)\label{fig:cm-dalma-msumg}]{
\includegraphics[width=0.4\linewidth]{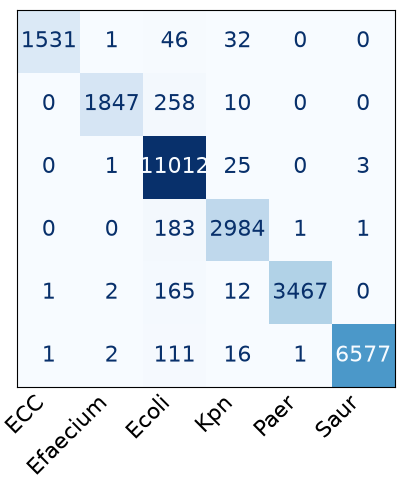}}
\caption{Confusion matrices illustrating the transferability of microbial identity across previously unseen clinical centers under the ZS protocol. 
}
\label{fig:confusion_matrices}
\end{figure}


\carlos{Figure~\ref{fig:confusion_matrices} provides a qualitative comparison between classifiers trained on the original spectral space and on DALMA latent representations. Classifiers trained on raw spectra show systematic cross-species confusions after deployment to unseen hospitals, most severely for \textit{Efaecium} on the MS-UMG cohort, whose spectra are acquired across two distinct instrument configurations (LT-SH and Smart) and share overlapping ribosomal-protein mass ranges with several gram-negative species, causing its raw-space signal to be dominated by acquisition artifacts rather than species identity. DALMA substantially reduces these errors by removing this acquisition-correlated variance, though it slightly overfits to \textit{Ecoli} on the MS-UMG cohort, likely because its large sample size still offers some batch-correlated cues the model can latch onto. These results indicate that the learned representation preserves microbial identity while reducing the presence of severe acquisition shifts.}



\begin{table}[!th]
\caption{OOD BA comparison across architectures.
Each model is used as a frozen feature extractor, and linear probing is applied on top of the resulting embeddings. Raw spectra reports performance of a fully supervised MLP trained on the original spectral space. Best result per row is highlighted in \textbf{bold}.}
\centering
\renewcommand{\arraystretch}{1.0}
\resizebox{\columnwidth}{!}{%
\begin{tabular}{llcccccccccc}
\toprule
\textbf{Train} & \textbf{Test} &
\textbf{Raw} &
\textbf{VAE} & \textbf{DANN} & \textbf{CORAL} & \textbf{MaldiT} &
\textbf{DALMA}\\
\midrule
\multirow{2}{*}{DRIAMS-A} & DRIAMS-D & 0.904 & 0.891 & 0.901 & 0.853 & 0.783 & \textbf{0.910} \\
                          & MS-UMG   & 0.950 & 0.723 & 0.945 & 0.519 & 0.167 & \textbf{0.952} \\
\midrule
\multirow{2}{*}{DRIAMS-B} & DRIAMS-D & 0.843 & 0.295 & 0.601 & 0.564 & 0.168 & \textbf{0.910} \\
                          & MS-UMG   & 0.860 & 0.345 & 0.649 & 0.458 & 0.161 & \textbf{0.949} \\
\midrule
\multirow{2}{*}{DRIAMS-C} & DRIAMS-D & 0.882 & 0.452 & 0.893 & 0.502 & 0.294 & \textbf{0.909} \\
                          & MS-UMG   & 0.928 & 0.408 & 0.947 & 0.330 & 0.158 & \textbf{0.948} \\
\midrule
\multirow{2}{*}{MARISMa}  & DRIAMS-D & 0.899 & 0.786 & 0.830 & 0.580 & 0.570 & \textbf{0.906} \\
                          & MS-UMG   & 0.884 & 0.331 & 0.903 & 0.175 & 0.213 & \textbf{0.945} \\
\midrule
\multirow{2}{*}{RKI}      & DRIAMS-D & 0.663 & 0.396 & 0.557 & 0.357 & 0.204 & \textbf{0.903} \\
                          & MS-UMG   & 0.732 & 0.220 & 0.547 & 0.241 & 0.152 & \textbf{0.954} \\
\midrule
\multirow{2}{*}{All}      & DRIAMS-D & \textbf{0.911} & 0.894 & 0.902 & 0.821 & 0.841 & \textbf{0.911} \\
                          & MS-UMG   & 0.870           & 0.511  & \textbf{0.949}  & 0.240  & 0.224  & \textbf{0.949} \\
\bottomrule
\end{tabular}%
}
\label{tab:ood_comparison}
\end{table}

Table~\ref{tab:ood_comparison} provides a quantitative comparison across all source-target configurations. DALMA consistently achieves the highest BA across nearly all experimental settings, maintaining performance above 0.90 for both unseen target cohorts, irrespective of the acquisition center used for training.

The improvements are particularly remarkable for challenging source domains such as DRIAMS-B and RKI, where conventional VAEs and domain adaptation methods exhibit severe performance degradation. Even large pretrained MALDI foundation models fail to generalize consistently across hospitals, whereas DALMA remains remarkably stable. These results demonstrate that explicitly organizing the latent representation around microbial identity yields substantially more transferable representations than existing representation learning or domain adaptation approaches.

The strong ZS performance also leaves little room for improvement through FS adaptation. As reported in the Appendix \ref{app:fewshot}
, fine-tuning with a small number of labeled target samples produces only marginal performance gains, indicating that the learned latent representation has already captured most transferable biological information.


\subsubsection{Ablation Study}
\label{subsec:ablation}

To better understand the origin of these improvements, here we analyze the contribution of the two architectural components responsible for learning biologically transferable representations: the species-conditioned latent prior and the domain-specific decoder architecture.

\begin{table}[!th]
\caption{Architectural analysis of DALMA, isolating the contribution of the species-conditioned latent prior (standard vs. conditional) and the domain-specific decoder architecture to cross-center transfer (single vs. multi-decoder). Best result in each row is highlighted in \textbf{bold}.}
\centering
\renewcommand{\arraystretch}{1.0}
\resizebox{\columnwidth}{!}{%
\begin{tabular}{ll cc cc}
\toprule
\multirow{2}{*}{\textbf{Train}} & \multirow{2}{*}{\textbf{Test}} &
\multicolumn{2}{c}{\textbf{Shared decoder}} &
\multicolumn{2}{c}{\textbf{Multi-decoder}} \\
\cmidrule(lr){3-4} \cmidrule(lr){5-6}
& & Standard & Conditioned & Standard & Conditioned \\
\midrule
\multirow{2}{*}{DRIAMS-A} & DRIAMS-D & 0.891 & \textbf{0.912} & 0.900 & 0.910 \\
         & MS-UMG   & 0.723 & 0.950 & 0.562 & \textbf{0.952} \\
\midrule
\multirow{2}{*}{DRIAMS-B} & DRIAMS-D & 0.295 & \textbf{0.910} & 0.386 & \textbf{0.910} \\
         & MS-UMG   & 0.345 & \textbf{0.949} & 0.300 & \textbf{0.949} \\
\midrule
\multirow{2}{*}{DRIAMS-C} & DRIAMS-D & 0.452 & \textbf{0.912} & 0.421 & \textbf{0.909} \\
         & MS-UMG   & 0.408 & \textbf{0.950} & 0.212 & 0.948 \\
\midrule
\multirow{2}{*}{MARISMa}  & DRIAMS-D & 0.786 & \textbf{0.913} & 0.720 & 0.906 \\
         & MS-UMG   & 0.331 & 0.939 & 0.591 & \textbf{0.945} \\
\midrule
\multirow{2}{*}{RKI}      & DRIAMS-D & 0.396 & 0.731 & 0.282 & \textbf{0.903} \\
         & MS-UMG   & 0.220 & 0.952 & 0.170 & \textbf{0.954} \\
\midrule
\multirow{2}{*}{All}      & DRIAMS-D & 0.894 & 0.910 & 0.900 & \textbf{0.911} \\
         & MS-UMG   & 0.511 & 0.938 & 0.618 & \textbf{0.949} \\
\bottomrule
\end{tabular}%
}
\label{tab:ablation}
\end{table}

Table~\ref{tab:ablation} compares the four possible combinations obtained by replacing the species-conditioned prior with a standard isotropic Gaussian prior and the domain-specific decoders with a single shared decoder.

The conditioned latent prior consistently provides the largest performance improvement across all source-target configurations. Replacing the standard Gaussian prior with species-conditioned distributions dramatically increases BA, confirming that explicitly organizing the latent representation according to microbial identity is the primary source of the observed transferability.

The domain-specific decoder provides a complementary improvement by further reducing acquisition-specific variability during reconstruction. Although the conditioned prior alone already produces highly transferable representations, combining it with acquisition-specific decoders consistently eliminates the remaining failure cases, yielding the most robust performance across all experimental settings.


\subsection{AMR Auxiliary Supervision}
\label{subsec:amr_supervision}

We next evaluate the second biological supervision mechanism proposed in DALMA. Unlike microbial identification, AMR prediction cannot be naturally incorporated through a species-conditioned latent prior, since each isolate is simultaneously associated with multiple antibiotic-specific resistance labels. Instead, DALMA injects biological information into the latent representation through the auxiliary AMR prediction head introduced in Section \ref{subsec:biological_supervision}.

Following the experimental protocol described in Section \ref{subsec:amr_protocol}, the evaluation is focused on \textit{Kpn.} 
We evaluate the learned representations under both ZS and FS transfer settings, assessing whether auxiliary biological supervision successfully captures transferable phenotypic information across previously unseen acquisition domains.

\begin{figure*}[!th]
\centering
\includegraphics[width=\linewidth]{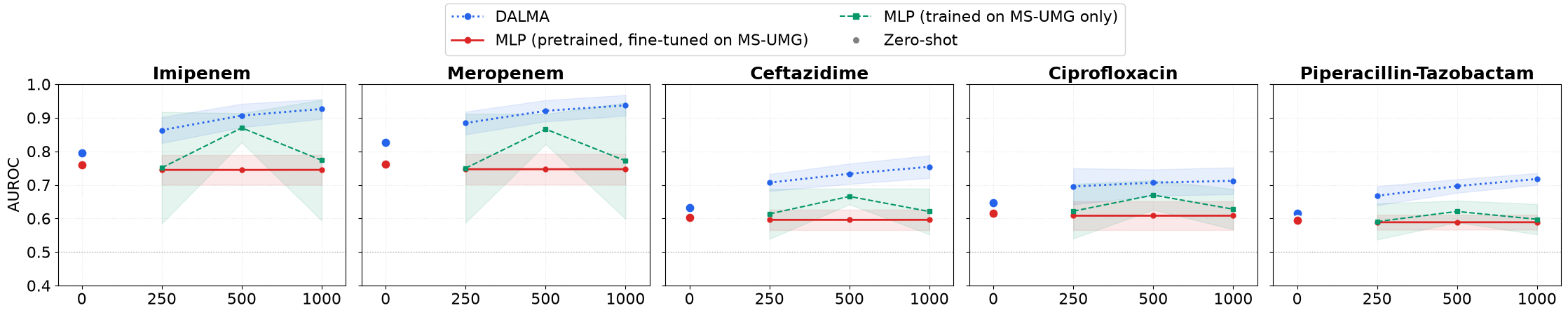}
\caption{ZS and FS transfer performance for AMR prediction in \textit{Kpn}. Results are shown for five clinically relevant antibiotics as a function of the number of labeled target-domain samples used for FS (x-axis), with 0 denoting the ZS setting. 
Shaded regions denote the standard deviation across 10 random partitions.
}
\label{fig:amr_transfer}
\end{figure*}

Figure~\ref{fig:amr_transfer} summarizes AMR prediction performance for the five antibiotics evaluated in \textit{Kpn}. The leftmost point in each subplot corresponds to the ZS setting, where models are directly transferred to the target acquisition domain without observing any labeled target samples. The remaining points report FS adaptation as progressively larger numbers of labeled isolates become available for fine-tuning.

Across all antibiotics, DALMA consistently achieves the strongest ZS performance, demonstrating that auxiliary biological supervision successfully incorporates AMR information into the learned latent representation despite never observing target-domain labels during training.

When limited labeled data are available, DALMA rapidly improves its predictive performance while maintaining a clear advantage over both pretrained and task-specific baselines. This behavior indicates that the learned representations already capture general resistance-related patterns, requiring only minimal adaptation to accommodate acquisition characteristics.

The magnitude of these improvements depends on the intrinsic difficulty of each resistance phenotype. For carbapenems, such as Imipenem and Meropenem, ZS performance is already remarkably high, leaving little room for additional improvement through fine-tuning. In contrast, more challenging antibiotics, including Ceftazidime and Ciprofloxacin, benefit more substantially from the incorporation of labeled target samples while preserving the relative advantage of DALMA over competing approaches.

Overall, these experiments demonstrate that the proposed auxiliary supervision strategy effectively extends biologically informed representation learning beyond taxonomic identity. Whereas microbial identification naturally enables species-conditioned latent priors, AMR can instead be incorporated through auxiliary supervision, allowing DALMA to learn transferable representations that capture both acquisition-invariant and clinically relevant phenotypic information.

\subsection{Robustness to Unseen Domain Shifts}
\label{subsec:ood}

The previous experiments demonstrate that DALMA learns latent representations that are largely invariant to the acquisition domains observed during training. However, real clinical deployment inevitably involves previously unseen sources of variability that cannot be fully anticipated during representation learning. Rather than assuming complete robustness to every possible domain shift, an equally important requirement is the ability to recognize when an incoming sample falls outside the learned distribution. We therefore evaluate whether the latent representations learned by DALMA support reliable OOD detection.

Following the protocol described in Section \ref{subsec:experimental_protocols}, GMMs are fitted to the latent representations learned from the source acquisition domains. OOD detection is then evaluated on spectra from previously unseen clinical centers by using the latent likelihood as a novelty score for selective prediction.

\begin{table}[th]
\caption{Selective prediction under domain shift. BA after progressively rejecting spectra with the lowest latent likelihood (highest novelty). Coverage denotes the percentage of retained samples.}
\centering
\renewcommand{\arraystretch}{1.3}
\resizebox{\columnwidth}{!}{%
\begin{tabular}{lcccc}
\toprule
\textbf{Set} & \textbf{Percentile} & \textbf{Discarded / Total} & \textbf{Coverage (\%)} & \textbf{BA} \\
\midrule
\multirow{5}{*}{DRIAMS-D}    & No rejection & 0 / 7,269    & 100.0  & 0.911 \\
& 0.01 & 12 / 7,269    & 99.8  & 0.911 \\
                              & 0.05 & 92 / 7,269    & 98.7  & 0.918 \\
                              & 0.10 & 401 / 7,269   & 94.5  & 0.941 \\
                              & 0.50 & 738 / 7,269   & 89.8  & \textbf{0.977} \\
\midrule
\multirow{5}{*}{MS-UMG} & No rejection & 0 / 28,290    & 100.0  & 0.949 \\
& 0.01 & 0 / 28,290      & 100.0 & 0.950 \\
                              & 0.05 & 1,147 / 28,290  & 95.9  & 0.992 \\
                              & 0.10 & 3,476 / 28,290  & 87.7  & \textbf{0.997} \\
                              & 0.50 & 20,940 / 28,290 & 26.0  & 0.996 \\
\bottomrule
\end{tabular}%
}
\label{tab:novelty_coverage}
\end{table}

Table~\ref{tab:novelty_coverage} summarizes the trade-off between coverage and BA obtained by progressively rejecting spectra with the lowest latent likelihood.  For both target domains, discarding only a small fraction of the most atypical samples consistently improves classification performance while maintaining high coverage. The 0.10 percentile threshold offers a favorable operating point, achieving substantial BA improvement while retaining more than 94\% and 87\% of samples in DRIAMS-D and MS-UMG, respectively. These results indicate that the learned latent representation provides a meaningful estimate of sample novelty, allowing DALMA to identify potentially unreliable predictions arising from previously unseen domain shifts rather than forcing confident predictions for every incoming sample.


%% file: Inputs/conclusions_jbhi.tex


\carlos{We presented DALMA, a probabilistic framework for learning biologically informed latent representations from MALDI-TOF spectra. Rather than pursuing domain invariance as an objective in itself, DALMA structures the latent space around biological characteristics while explicitly modeling acquisition-specific variability through domain-specific decoders, allowing technical differences between laboratories to be absorbed without distorting the shared representation. Building on this domain-corrected representation, the species-conditioned latent prior emerges as the principal driver of biological transferability, enabling generalization across heterogeneous clinical centers without requiring target-domain adaptation. This disentanglement of technical and biological sources of variation provides a more effective strategy than enforcing complete domain invariance through adversarial or distribution-alignment objectives.}

\carlos{Beyond microbial identification, the learned representations transferred clinically relevant phenotypic information to downstream AMR prediction through an auxiliary supervision mechanism, achieving strong zero-shot performance and rapid adaptation from limited labeled target samples — indicating that the representations capture both conserved biological signatures and task-specific phenotypic information. 

As complete robustness to every possible distribution shift cannot be guaranteed, the proposed novelty detection framework further allows the model to recognize spectra falling outside the learned latent distribution and abstain from unreliable predictions, providing an additional layer of safety for clinical deployment. 

More broadly, our results suggest that biological supervision offers a principled alternative to purely domain-invariant representation learning, a paradigm that may extend beyond MALDI-TOF MS to other biomedical problems where biological variability and acquisition heterogeneity coexist.}

%% file: Appendix.tex
\section{Baseline Architectures and Training Details}
\label{app:baselines}

This section provides architectural and training details for the baseline models used in the comparative evaluation. DALMA's architecture and training procedure are described in Sections\ref{sec:methodology} and \ref{subsec:implementation-details} of the main paper.

\subsection{Raw Spectra Baseline}
The fully supervised MLP trained directly on the raw spectra mirrors the hidden-layer structure of DALMA's encoder, with hidden layers of 2,048, 1,024, and 512 units (ReLU activations), followed by an additional 64-unit ReLU layer matching DALMA's latent dimensionality, and a final linear classification layer with $S$ output units (one per target species), trained with cross-entropy loss. This ensures that differences in performance relative to DALMA reflect the value of the learned latent representation itself, rather than differences in raw model capacity.

\subsection{Standard VAE}
The standard VAE baseline uses the same encoder and decoder architecture as DALMA, with a fixed isotropic Gaussian prior $p(z) = \mathcal{N}(0, I)$ instead of the species-conditioned prior. Architectural details are summarized in Tables~\ref{tab:vae_encoder} and~\ref{tab:vae_decoder}.

\begin{table}[H]
\centering
\caption{VAE Encoder Architecture.}
\label{tab:vae_encoder}
\begin{tabular}{lll}
\hline
\textbf{Layer} & \textbf{Configuration} & \textbf{Activation} \\
\hline
Input & 6,000 m/z bins & -- \\
Fully Connected 1 & 2,048 units & ReLU \\
Fully Connected 2 & 1,024 units & ReLU \\
Fully Connected 3 & 512 units & ReLU \\
Latent Mean ($\mu$) & 64 units & -- \\
Latent Log-variance ($\log \sigma^2$) & 64 units & Clamp $(-6, 6)$ \\
\hline
\end{tabular}%

\end{table}

\begin{table}[H]
\centering
\caption{VAE Decoder Architecture.}
\label{tab:vae_decoder}
\begin{tabular}{lll}
\hline
\textbf{Layer} & \textbf{Configuration} & \textbf{Activation} \\
\hline
Input & Latent $z$ (64 units) & -- \\
Fully Connected 1 & 512 units & ReLU \\
Fully Connected 2 & 1,024 units & ReLU \\
Fully Connected 3 & 2,048 units & ReLU \\
Output Layer & 6,000 units & Sigmoid \\
\hline
\end{tabular}
\end{table}

\subsection{Domain-Adversarial Neural Network (DANN)}
DANN is implemented following the original formulation in~\cite{ganin2016}, comprising a feature extractor ($G_f$), a label predictor ($G_y$), and a domain discriminator ($G_d$). A gradient reversal layer is inserted between $G_f$ and $G_d$, reversing the gradient sign during backpropagation to encourage domain-invariant representations.

\begin{table}[H]
\centering
\caption{DANN Architecture Specification.}
\label{tab:dann_arch}
\begin{tabular}{lll}
\hline
\textbf{Module} & \textbf{Layer} & \textbf{Configuration} \\
\hline
\multirow{3}{*}{Feature Extractor ($G_f$)}
 & FC 1       & 1,024 units (ReLU) \\
 & FC 2       & 1,024 units (ReLU) \\
 & Latent $z$ & 64 units (Linear) \\
\hline
\multirow{3}{*}{Label Predictor ($G_y$)}
 & FC 3   & 1,024 units (ReLU) \\
 & FC 4   & 1,024 units (ReLU) \\
 & Output & $S$ species (Softmax) \\
\hline
\multirow{3}{*}{Discriminator ($G_d$)}
 & FC 5   & 1,024 units (ReLU) \\
 & FC 6   & 1,024 units (ReLU) \\
 & Output & $D$ domains (Softmax) \\
\hline
\end{tabular}%
\end{table}

\subsection{MultiVAE-CORAL}
MultiVAE-CORAL extends the multi-decoder VAE framework with a correlation alignment regularizer. It shares the same encoder and decoder architecture as DALMA. During training, the ELBO is augmented with a multi-domain CORAL loss that minimizes the Frobenius distance between covariance matrices of all source-domain pairs in the latent space:

\begin{equation}
C_k = \frac{1}{n_k - 1} (Z_k - \bar{Z}_k)^\top (Z_k - \bar{Z}_k),
\end{equation}

\begin{equation}
\mathcal{L}_{\mathrm{CORAL}} = \frac{1}{P} \sum_{i=1}^{K} \sum_{j > i}^{K} \frac{1}{4d^2} \|C_i - C_j\|^2_F,
\end{equation}

where $P$ is the number of domain pairs, $d = 64$ is the latent dimensionality, and the final objective is $\mathcal{L} = \mathcal{L}_{\mathrm{ELBO}} + \lambda_{\mathrm{CORAL}} \mathcal{L}_{\mathrm{CORAL}}$. Hyperparameters are summarized in Table~\ref{tab:coral_hparams}.

\begin{table}[H]
\centering
\caption{MultiVAE-CORAL Training Hyperparameters.}
\label{tab:coral_hparams}
\begin{tabular}{ll}
\hline
\textbf{Hyperparameter} & \textbf{Value} \\
\hline
Optimizer & Adam \\
Learning Rate & $10^{-4}$ \\
Weight Decay & $10^{-5}$ \\
$\lambda_{\mathrm{CORAL}}$ & $10^{-3}$ \\
Latent Dimension & 64 \\
Early Stopping Patience & 20 epochs \\
\hline
\end{tabular}
\end{table}

\subsection{Maldi Transformer}
The Maldi Transformer~\cite{dewaele2025pre} comparison uses the pretrained Extra-Large (XL) variant as a frozen feature extractor. Each input spectrum is restricted to the top-200 highest-intensity peaks, and the 512-dimensional \texttt{[CLS]} token embedding is extracted without any fine-tuning and used as input to the downstream linear classifier.

\subsection{Training Hyperparameters}
Optimization was performed using the Adam optimizer~\cite{kingma2014} for all models except DANN, which uses SGD. Hyperparameters are summarized in Table~\ref{tab:training_hparams}.

\begin{table}[H]
\centering
\caption{Optimization and Training Hyperparameters.}
\label{tab:training_hparams}
\begin{tabular}{lcc}
\hline
\textbf{Hyperparameter} & \textbf{VAE-based} & \textbf{DANN} \\
\hline
Optimizer & Adam & SGD (mom: 0.9) \\
Learning Rate & $10^{-4}$ & $10^{-5}$ \\
Weight Decay & $10^{-5}$ & -- \\
Gradient Clipping & -- & 1.0 \\
Batch Size & 128 & 128 \\
Max Epochs & 100 & 100 \\
Early Stopping Patience & 20 & 20 \\
\hline
\end{tabular}
\end{table}

\section{Extended Evaluation Results}
\label{app:ood_extended}
This section complements the main results of Section V.B with the full set of evaluation metrics across all architectures and out-of-distribution targets. Table~\ref{tab:ood_f1} reports macro-averaged F1, Table~\ref{tab:ood_recall} macro-averaged recall, Table~\ref{tab:ood_specificity} macro-averaged specificity, and Table~\ref{tab:ood_auroc} AUROC. 

Across all four metrics, the pattern observed for balanced accuracy in Section V.B is largely reproduced. DALMA achieves the highest or near-highest F1-macro and recall in nearly every source-target combination, with the largest differences on the smallest source domains (DRIAMS-B, RKI). Specificity is uniformly high across all methods, reflecting its lower sensitivity to minority-species misclassification. AUROC shows a more mixed picture: DALMA remains strongest on DRIAMS-D, but the raw spectral baseline is occasionally competitive on MS-UMG, suggesting that ranking-based metrics can mask the larger gaps seen in hard classification performance.

\begin{table}[!t]
\caption{OOD F1-macro comparison across architectures. Best result per row in \textbf{bold}.}
\centering
\renewcommand{\arraystretch}{1.0}
\resizebox{\columnwidth}{!}{%
\begin{tabular}{ll cccccc}
\toprule
\textbf{Train} & \textbf{Test} & \textbf{Raw} & \textbf{VAE} & \textbf{DANN} & \textbf{CORAL} & \textbf{MaldiT-XL} & \textbf{DALMA} \\
\midrule
\multirow{2}{*}{DRIAMS-A} & DRIAMS-D & 0.842 & \textbf{0.876} & 0.842 & 0.815 & 0.708 & 0.867 \\
         & MS-UMG   & 0.961 & 0.717 & 0.939 & 0.500 & 0.055 & \textbf{0.967} \\
\midrule
\multirow{2}{*}{DRIAMS-B} & DRIAMS-D & 0.832 & 0.257 & 0.617 & 0.443 & 0.088 & \textbf{0.874} \\
         & MS-UMG   & 0.891 & 0.290 & 0.606 & 0.466 & 0.039 & \textbf{0.965} \\
\midrule
\multirow{2}{*}{DRIAMS-C} & DRIAMS-D & 0.823 & 0.434 & 0.851 & 0.448 & 0.247 & \textbf{0.872} \\
         & MS-UMG   & 0.951 & 0.390 & 0.927 & 0.252 & 0.060 & \textbf{0.959} \\
\midrule
\multirow{2}{*}{MARISMa}  & DRIAMS-D & \textbf{0.871} & 0.734 & 0.791 & 0.565 & 0.567 & 0.853 \\
         & MS-UMG   & 0.869 & 0.241 & 0.892 & 0.060 & 0.093 & \textbf{0.961} \\
\midrule
\multirow{2}{*}{RKI}      & DRIAMS-D & 0.559 & 0.183 & 0.419 & 0.201 & 0.032 & \textbf{0.840} \\
         & MS-UMG   & 0.694 & 0.111 & 0.474 & 0.126 & 0.027 & \textbf{0.969} \\
\midrule
\multirow{2}{*}{All}      & DRIAMS-D & 0.838 & \textbf{0.884} & 0.834 & 0.838 & 0.809 & 0.860 \\
         & MS-UMG   & 0.839 & 0.467 & 0.945 & 0.170 & 0.106 & \textbf{0.965} \\
\bottomrule
\end{tabular}%
}
\label{tab:ood_f1}
\end{table}

\begin{table}[!t]
\caption{OOD macro-averaged recall comparison across architectures. Best result per row in \textbf{bold}.}
\centering
\renewcommand{\arraystretch}{1.0}
\resizebox{\columnwidth}{!}{%
\begin{tabular}{ll cccccc}
\toprule
\textbf{Train} & \textbf{Test} & \textbf{Raw} & \textbf{VAE} & \textbf{DANN} & \textbf{CORAL} & \textbf{MaldiT-XL} & \textbf{DALMA} \\
\midrule
\multirow{2}{*}{DRIAMS-A} & DRIAMS-D & 0.904 & 0.891 & 0.901 & 0.853 & 0.783 & \textbf{0.910} \\
         & MS-UMG   & 0.950 & 0.723 & 0.945 & 0.519 & 0.167 & \textbf{0.952} \\
\midrule
\multirow{2}{*}{DRIAMS-B} & DRIAMS-D & 0.843 & 0.295 & 0.601 & 0.564 & 0.168 & \textbf{0.910} \\
         & MS-UMG   & 0.860 & 0.345 & 0.649 & 0.458 & 0.161 & \textbf{0.949} \\
\midrule
\multirow{2}{*}{DRIAMS-C} & DRIAMS-D & 0.882 & 0.452 & 0.893 & 0.502 & 0.294 & \textbf{0.909} \\
         & MS-UMG   & 0.928 & 0.408 & 0.947 & 0.330 & 0.158 & \textbf{0.948} \\
\midrule
\multirow{2}{*}{MARISMa}  & DRIAMS-D & 0.899 & 0.786 & 0.830 & 0.580 & 0.570 & \textbf{0.906} \\
         & MS-UMG   & 0.884 & 0.331 & 0.903 & 0.175 & 0.213 & \textbf{0.945} \\
\midrule
\multirow{2}{*}{RKI}      & DRIAMS-D & 0.663 & 0.396 & 0.557 & 0.357 & 0.204 & \textbf{0.903} \\
         & MS-UMG   & 0.732 & 0.220 & 0.547 & 0.241 & 0.152 & \textbf{0.954} \\
\midrule
\multirow{2}{*}{All}      & DRIAMS-D & \textbf{0.911} & 0.894 & 0.902 & 0.821 & 0.841 & \textbf{0.911} \\
         & MS-UMG   & 0.870 & 0.511 & 0.949 & 0.240 & 0.224 & \textbf{0.949} \\
\bottomrule
\end{tabular}%
}
\label{tab:ood_recall}
\end{table}

\begin{table}[!t]
\caption{OOD macro-averaged specificity comparison across architectures. Best result per row in \textbf{bold}.}
\centering
\renewcommand{\arraystretch}{1.0}
\resizebox{\columnwidth}{!}{%
\begin{tabular}{ll cccccc}
\toprule
\textbf{Train} & \textbf{Test} & \textbf{Raw} & \textbf{VAE} & \textbf{DANN} & \textbf{CORAL} & \textbf{MaldiT-XL} & \textbf{DALMA} \\
\midrule
\multirow{2}{*}{DRIAMS-A} & DRIAMS-D & 0.979 & \textbf{0.980} & \textbf{0.980} & 0.965 & 0.959 & 0.979 \\
         & MS-UMG   & 0.991 & 0.968 & \textbf{0.992} & 0.937 & 0.833 & \textbf{0.992} \\
\midrule
\multirow{2}{*}{DRIAMS-B} & DRIAMS-D & 0.959 & 0.867 & 0.959 & 0.918 & 0.834 & \textbf{0.980} \\
         & MS-UMG   & 0.979 & 0.907 & 0.967 & 0.927 & 0.832 & \textbf{0.992} \\
\midrule
\multirow{2}{*}{DRIAMS-C} & DRIAMS-D & 0.979 & 0.900 & \textbf{0.980} & 0.924 & 0.861 & 0.979 \\
         & MS-UMG   & 0.990 & 0.920 & \textbf{0.991} & 0.908 & 0.830 & \textbf{0.991} \\
\midrule
\multirow{2}{*}{MARISMa}  & DRIAMS-D & 0.977 & 0.956 & 0.972 & 0.910 & 0.933 & \textbf{0.979} \\
         & MS-UMG   & 0.985 & 0.858 & 0.988 & 0.835 & 0.838 & \textbf{0.992} \\
\midrule
\multirow{2}{*}{RKI}      & DRIAMS-D & 0.922 & 0.856 & 0.902 & 0.875 & 0.835 & \textbf{0.979} \\
         & MS-UMG   & 0.955 & 0.841 & 0.936 & 0.849 & 0.832 & \textbf{0.992} \\
\midrule
\multirow{2}{*}{All}      & DRIAMS-D & \textbf{0.980} & 0.976 & 0.980 & 0.966 & 0.974 & \textbf{0.980} \\
         & MS-UMG   & 0.973 & 0.902 & \textbf{0.992} & 0.852 & 0.840 & 0.991 \\
\bottomrule
\end{tabular}%
}
\label{tab:ood_specificity}
\end{table}

\begin{table}[!t]
\caption{OOD AUROC comparison across architectures. Best result per row in \textbf{bold}.}
\centering
\renewcommand{\arraystretch}{1.0}
\resizebox{\columnwidth}{!}{%
\begin{tabular}{ll cccccc}
\toprule
\textbf{Train} & \textbf{Test} & \textbf{Raw} & \textbf{VAE} & \textbf{DANN} & \textbf{CORAL} & \textbf{MaldiT-XL} & \textbf{DALMA} \\
\midrule
\multirow{2}{*}{DRIAMS-A} & DRIAMS-D & 0.988 & 0.981 & 0.981 & 0.973 & 0.938 & \textbf{0.995} \\
         & MS-UMG   & \textbf{0.999} & 0.966 & 0.992 & 0.955 & 0.572 & 0.984 \\
\midrule
\multirow{2}{*}{DRIAMS-B} & DRIAMS-D & 0.966 & 0.807 & 0.967 & 0.873 & 0.580 & \textbf{0.990} \\
         & MS-UMG   & \textbf{0.994} & 0.874 & 0.973 & 0.929 & 0.563 & 0.984 \\
\midrule
\multirow{2}{*}{DRIAMS-C} & DRIAMS-D & 0.987 & 0.933 & 0.972 & 0.935 & 0.785 & \textbf{0.995} \\
         & MS-UMG   & \textbf{0.998} & 0.880 & 0.986 & 0.936 & 0.505 & 0.989 \\
\midrule
\multirow{2}{*}{MARISMa}  & DRIAMS-D & 0.987 & 0.963 & 0.985 & 0.922 & 0.899 & \textbf{0.995} \\
         & MS-UMG   & \textbf{0.992} & 0.924 & 0.993 & 0.905 & 0.635 & 0.975 \\
\midrule
\multirow{2}{*}{RKI}      & DRIAMS-D & 0.916 & 0.822 & 0.860 & 0.822 & 0.594 & \textbf{0.971} \\
         & MS-UMG   & 0.979 & 0.795 & 0.868 & 0.798 & 0.532 & \textbf{0.984} \\
\midrule
\multirow{2}{*}{All}      & DRIAMS-D & 0.995 & 0.986 & 0.989 & 0.985 & 0.967 & \textbf{0.996} \\
         & MS-UMG   & 0.975 & 0.971 & 0.994 & 0.965 & 0.629 & \textbf{0.989} \\
\bottomrule
\end{tabular}%
}
\label{tab:ood_auroc}
\end{table}

\section{Few-Shot Domain Adaptation}
\label{app:fewshot}

\begin{figure}[!th]
    \centering
    \includegraphics[width=\linewidth]{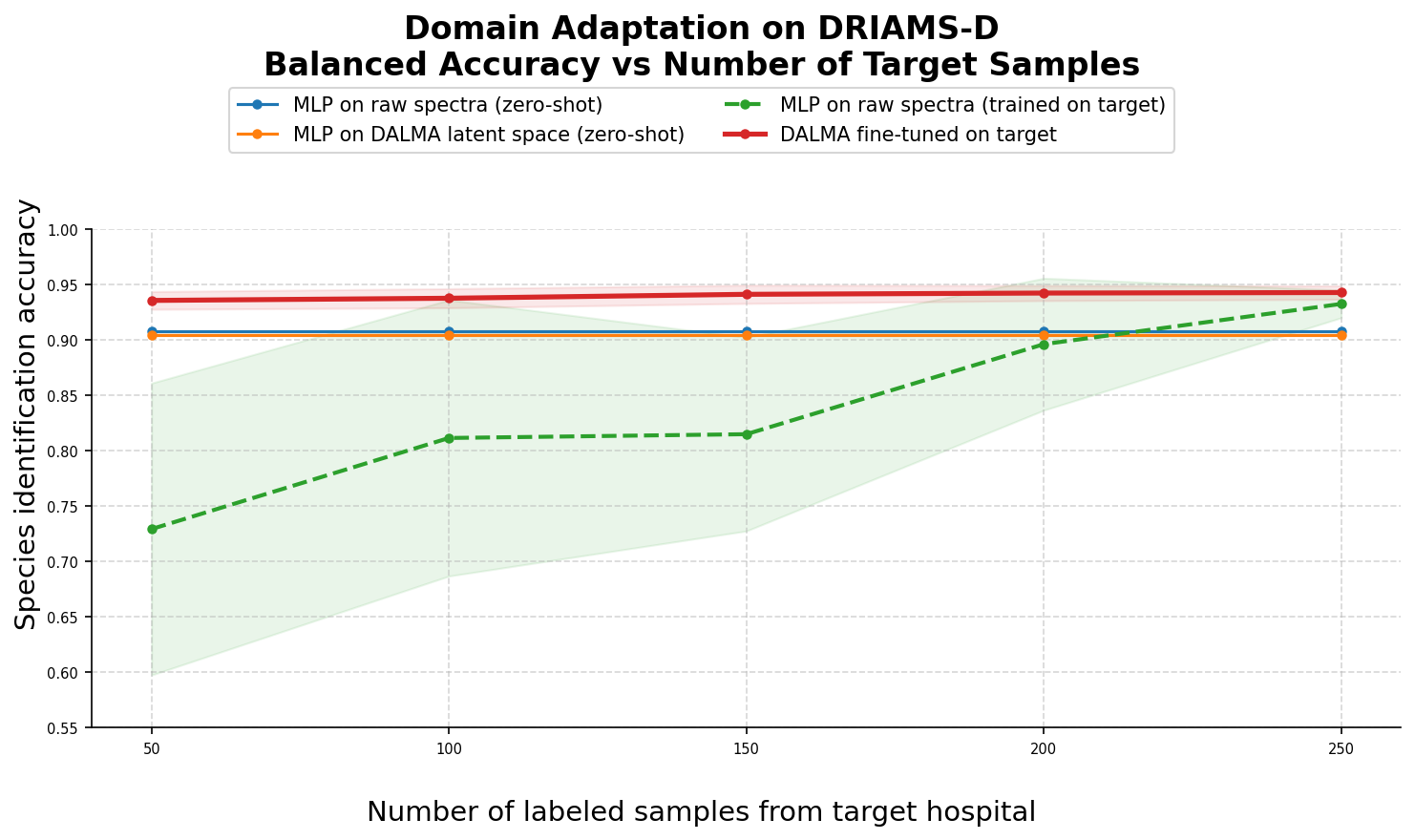}
    \caption{Few-shot domain adaptation performance on DRIAMS-D. Shaded areas represent standard deviation across 10 random partitions.}
    \label{fig:grid_driams_d}
\end{figure}

\begin{figure}[!th]
    \centering
    \includegraphics[width=\linewidth]{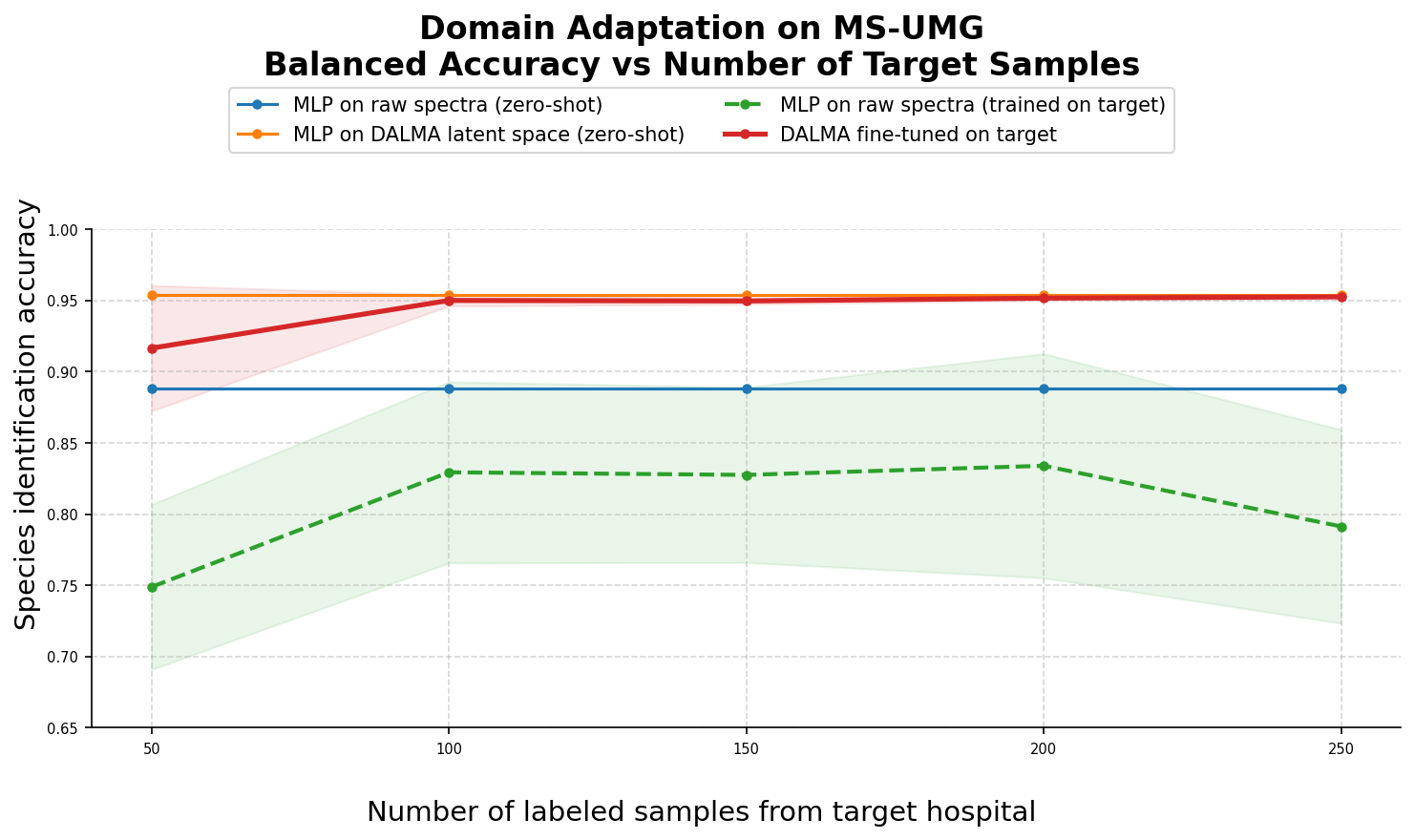}
    \caption{Few-shot domain adaptation performance on MS-UMG. Shaded areas represent standard deviation across 10 random partitions.}
    \label{fig:grid_ms_umg}
\end{figure}

This section complements Section V.B by reporting the full few-shot domain adaptation experiments.

On DRIAMS-D (Figure~\ref{fig:grid_driams_d}), both zero-shot baselines stabilize around $0.905$ regardless of sample size. A target-only classifier trained from scratch starts at $0.73$ with 50 samples and needs 250 samples to reach $0.93$, while fine-tuning DALMA already reaches $0.938$ with only 50 samples.

On MS-UMG (Figure~\ref{fig:grid_ms_umg}), DALMA in zero-shot ($0.954$) already outperforms the raw spectral MLP ($0.870$), the opposite trend observed on DRIAMS-D. Training from scratch is unstable, peaking near $0.83$ before dropping to $0.79$ at 250 samples. Fine-tuning DALMA starts at $0.914$ with 50 samples and converges to $0.952$ by 100 samples.